\pdfoutput=1
\documentclass[11pt]{article}

\usepackage[margin=0.75in]{geometry}
\usepackage{enumitem}

\usepackage[table]{xcolor}
\definecolor{lightgreen}{RGB}{198, 239, 206}
\definecolor{firstcol}{rgb}{1.00,0.80,0.80}   
\definecolor{secondcol}{rgb}{1.00,0.93,0.70}  

\definecolor{sirenblue}{RGB}{31,119,180}
\definecolor{finerred}{RGB}{214,39,40}

\usepackage{graphicx}
\usepackage{amsmath}
\usepackage{amssymb}
\usepackage{amsthm} 
\usepackage{authblk} 

\usepackage{float}
\usepackage{booktabs}
\usepackage[pagebackref,breaklinks,colorlinks]{hyperref}
\usepackage[capitalize]{cleveref}

\newtheorem{proposition}{Proposition}
\theoremstyle{remark}
\newtheorem{remark}{Remark}

\usepackage{pgfplots}
\pgfplotsset{compat=1.17}
\usepgfplotslibrary{groupplots}
\usepackage{tikz}

\usepackage{multirow}
\crefname{section}{Sec.}{Secs.}
\Crefname{section}{Section}{Sections}
\Crefname{table}{Table}{Tables}
\crefname{table}{Tab.}{Tabs.}

\title{FiRe: Frequency Reparameterization as a Preconditioner for Periodic Implicit Neural Representations}

\author[1]{Harinandan Shuka}
\author[1]{Rajarshi Verma}
\author[2,*]{Jitin Singla}

\affil[1]{Department of Computer Science and Engineering, IIT Roorkee}
\affil[2]{Department of Biosciences and Bioengineering, IIT Roorkee}
\affil[*]{Corresponding Author}

\date{}

\begin{document}

\maketitle

\begin{abstract}
Periodic Implicit Neural Representations (INRs) such as SIREN and FINER assign every neuron, the same global frequency $\omega_0$, spending the representational budget inefficiently when local signal content varies. We introduce \textbf{FiRe} (\emph{Frequency Reparameterization}), that accelerates optimization by reparameterizing per-neuron frequency of periodic INRs without changing their underlying activation function. FiRe gives each neuron a bounded, input-dependent frequency via a separate low-rank gating path and is applicable to any periodic activation function. The gate acts as an implicit \textbf{preconditioner} that improves optimization conditioning at initialization via the Neural Tangent Kernel (NTK). This better-conditioned initialization makes optimization converge faster, and the high-frequency content of the reconstruction tracks the target more closely at a fixed computational budget. On 2D image fitting, FiRe increases PSNR over a parameter-matched baseline (up to $\approx\!+1$\,dB at short training budgets), with gains that vary with resolution and diminish at full convergence. We characterize how performance depends on resolution, rank, and training budget, and give an NTK account that predicts these trends.
\end{abstract}

\section{Introduction}
\label{sec:intro}
Implicit Neural Representations (INRs) encode a continuous signal as the weights of a coordinate network $f_\theta:\mathbb{R}^d\!\to\!\mathbb{R}^c$~\cite{sitzmann2020implicit,sitzmann2019scene}. The central obstacle is spectral bias~\cite{rahaman2019spectral}, i.e., gradient descent fits low frequencies first, and standard ReLU MLPs recover fine detail only slowly because the high-frequency eigenvalues of their tangent kernel are vanishingly small~\cite{jacot2018neural}. SIREN~\cite{sitzmann2020implicit} addressed this by replacing ReLU with sinusoidal activation function $\sin(\omega_0(\mathbf{W}\mathbf{x}+\mathbf{b}))$ under a variance-preserving initialization, and FINER~\cite{liu2024finer} extended it with a variable-periodic activation $\sin(\omega_0(|\mathbf{W}\mathbf{x}+\mathbf{b}|+1)(\mathbf{W}\mathbf{x}+\mathbf{b}))$ whose local frequency grows with the pre-activation magnitude.

This motivates making the per-neuron frequency \emph{learnable} and \emph{input/coordinate-dependent}. A neuron resolving a sharp edge need not oscillate at the same rate as one covering smooth background. We propose \textbf{FiRe} (\emph{Frequency Reparameterization}), a low-rank gate that gives each neuron a bounded, input-dependent frequency scale. Rather than folding the per-neuron frequency into a single weight matrix, FiRe re-expresses it in a factored, overparameterized form via a separate gate path that multiplies the weights (\S\ref{sec:method}). The realized frequency can be kept identical to that of a plain network; what changes is how it is parameterized.

The key question is what FiRe changes and why it helps. Two confounders can raise INR fitting quality independently of any reparameterization: a larger global frequency multiplier $\omega_0$ and a larger parameter count. We control for both. We match FiRe's realized per-neuron frequency to a plain baseline using a weight-rescaling control (\emph{wscale}; \S\ref{sec:control}), and we widen the baseline to match FiRe's parameter count. FiRe acts as an implicit preconditioner~\cite{salimans2016weight,arora2019implicit}, improving gradient-flow conditioning at initialization. Under matched realized frequency and matched parameters, FiRe's separate gate path changes the optimization geometry: it contributes an additional positive-semidefinite block to the Neural Tangent Kernel (NTK)~\cite{jacot2018neural} (\S\ref{sec:ntk}), enlarging and improving the conditioning of the tangent space at initialization. This yields faster early convergence (\S\ref{sec:long}) and gains that are concentrated in the high-frequency band (\S\ref{sec:qual},~\ref{sec:results-ntk}). A frequency-stability check confirms the gate keeps the baseline's realized frequency throughout training (\S\ref{sec:drift}). \\

\noindent\textbf{Contributions}
\begin{itemize}\itemsep2pt
  \item A per-neuron frequency gate (FiRe) for periodic INRs that routes frequency through a dedicated low-rank path, adds $1.4$--$7.7\%$ parameters, and composes with any periodic activation.
  \item An additive NTK decomposition into carrier and gate kernels: $K=K_{\text{c}}+K_{\text{g}}$ identifying FiRe as an implicit preconditioner; the gate path enlarges the tangent space ($+23$--$44\%$ effective rank, slower eigenvalue decay) at matched realized frequency, and we connect this enlargement to early-training convergence rate.
  \item A controlled study on 2D image fitting under a frequency-equalizing, parameter-matched comparison, establishing a seed-robust convergence-acceleration gain, a spectral analysis localizing it to the high-frequency band, and a map of how the gain depends on rank, resolution, and budget, matching the NTK account.
\end{itemize}

\section{Related Work}
The difficulty of fitting high-frequency signals with deep networks is explained by the Neural Tangent Kernel~\cite{jacot2018neural}, and the resulting spectral bias was characterized independently by Rahaman et al.~\cite{rahaman2019spectral}. A ReLU network's high-frequency tangent-kernel eigenvalues are vanishingly small, so it fits coarse structure first and recovers fine detail slowly. This bears directly on coordinate-based implicit representations, where ReLU MLPs encode signed distance functions~\cite{park2019deepsdf,chen2019learning}, occupancy fields~\cite{mescheder2019occupancy}, surfaces trained without 3D supervision~\cite{niemeyer2020differentiable}, Eikonal-regularized SDFs~\cite{gropp2020implicit}, and joint geometry-appearance scenes~\cite{sitzmann2019scene}; here spectral bias surfaces as blurred geometry and lost detail. Lifting coordinates through a Fourier feature mapping reshapes the kernel spectrum and unlocks high-frequency recovery~\cite{tancik2020fourier}, and the positional encoding of neural radiance fields~\cite{mildenhall2021nerf} applies the same idea to novel-view synthesis. Both, however, fix the frequency budget once, before training begins.

The dominant response is to abandon ReLU. Sinusoidal networks (SIREN)~\cite{sitzmann2020implicit} use $\sin(\omega_0(\mathbf{W}\mathbf{x}{+}\mathbf{b}))$ under a variance-preserving initialization, reshaping the NTK so that high-frequency modes become learnable; because every derivative of sine is again sinusoidal, derivative-level supervision such as Eikonal and Poisson constraints is natural. Gabor-wavelet~\cite{saragadam2023wire} and Gaussian~\cite{ramasinghe2022beyond} activations follow, trading frequency coverage for space-frequency localization and noise robustness, while multiplicative filter networks~\cite{fathony2020multiplicative} take element-wise products of sinusoidal or Gabor filters to span an exponentially large basis without stacking nonlinearities.

A second thread breaks the assumption that all neurons share one global frequency. Variable-periodic activations (FINER)~\cite{liu2024finer} feed the pre-activation magnitude back into the frequency envelope, producing a chirp whose effective frequency grows with magnitude. An NTK analysis shows a denser high-frequency eigenspectrum, explaining faster convergence. The principle extends to Gabor and Gaussian backbones~\cite{zhu2024finer++}, and a parallel line replaces the nonlinearity with learnable Chebyshev expansions that adapt their spectrum during training~\cite{rezaeian2025sl2a}. Frequency \emph{diversity} can instead be seeded at initialization: a Nyquist-informed, harmonically spaced per-neuron multiplier~\cite{alsakabi2025fm} halves hidden-feature redundancy at no parameter cost, but the multipliers are fixed and identical across spatial positions.

A step toward spatial adaptivity pairs frozen frequency embeddings with spatially-adaptive amplitude masks from a lightweight hash-grid~\cite{feng2026sasnet,muller2022instant}, localizing neuron influence while leaving oscillation frequency unchanged and adding non-trivial overhead. On the optimization-geometry side, reparameterization is known to reshape the gradient-descent trajectory without enlarging the function class~\cite{salimans2016weight,arora2019implicit}, and a Fourier reparameterization of the weight matrices rebalances the NTK spectrum and alleviates spectral bias at no inference cost~\cite{shi2024improved}. FiRe sits at the confluence of these threads. It applies a multiplicative, coordinate-dependent gate to per-neuron frequency through a dedicated low-rank path~\cite{hu2022lora}: per-neuron unlike sinusoidal and Gabor networks, coordinate-dependent unlike fixed harmonic multipliers, a continuous frequency modulation rather than an amplitude mask unlike spatial gating, and disjoint from the signal path unlike variable-periodic activations. Rather than assert an asymptotic reduction of spectral bias, we isolate the gate's effect through an exact additive NTK decomposition and identify it as a convergence-acceleration property at the short training budgets standard in INR fitting.

\section{Method}
\label{sec:method}
\subsection{Preliminaries}
INR encodes a continuous signal as the weights of a neural network rather than on a discrete grid. Given a signal defined over a coordinate domain $\Omega\subset\mathbb{R}^{d}$ ($d{=}2$ for images), an INR fits a multilayer perceptron $f_\theta:\mathbb{R}^{d}\to\mathbb{R}^{c}$ to a set of coordinate value pairs $\{(\mathbf{x}_i,\mathbf{y}_i)\}$, where $\mathbf{x}_i\in\mathbb{R}^d$ is an input coordinate
(e.g.\ a pixel location) and $\mathbf{y}_i\in\mathbb{R}^c$ is the corresponding signal value (e.g.\ an RGB colour). It minimizes MSE loss over $N$ training samples:
\begin{equation}
  \mathcal{L}(\theta)=
  \frac{1}{N}\sum_{i=1}^{N}
  \bigl\|f_\theta(\mathbf{x}_i)-\mathbf{y}_i\bigr\|^2,
  \label{eq:loss}
\end{equation}
Once fitted, the network can be queried at any coordinate, providing a continuous, resolution-independent representation of the signal.
\subsection{FiRe} As discussed above {SIREN}~\cite{sitzmann2020implicit} sets the activation of $f_\theta$ to $\mathbf{h}=\sin(\omega_0(\mathbf{W}\mathbf{x} + \mathbf{b}))$ with a global frequency multiplier $\omega_0$, with each neuron, in each layer. FiRe inserts a per-neuron, input-dependent gate $s(\mathbf{x})\in(0,2)^{d_h}$ that rescales the frequency before the activation:
\begin{equation}
  \mathbf{h} = \sin\!\bigl(
    \omega_0 \cdot s(\mathbf{x}) \odot
    (\mathbf{W}\mathbf{x} + \mathbf{b})
  \bigr),
  \label{eq:fire}
\end{equation}
where the gate is implemented as a low-rank bottleneck,
\begin{equation}
  s(\mathbf{x}) = 2\,\sigma\!\bigl(
    \mathbf{U}_{\uparrow}\mathbf{U}_{\downarrow}\mathbf{x}
    + \mathbf{b}_{\uparrow}
  \bigr),
  \label{eq:gate}
\end{equation}
with $\mathbf{U}_\uparrow\!\in\! \mathbb{R}^{d_h\times r}$, $\mathbf{U}_\downarrow\!\in\!\mathbb{R}^{r\times d}$, $\mathbf{b}_{\uparrow}\!\in\!\mathbb{R}^{d_h}$, rank $r$, and $\sigma$ the logistic sigmoid. Neuron $i$ operates at effective frequency $\omega_0 \mathbf{s}_i\in(0,2\omega_0)$. The gate path $(\mathbf{U}_\downarrow,\mathbf{U}_\uparrow,\mathbf{b}_\uparrow)$ is separate from the signal path $(\mathbf{W},\mathbf{b})$, and the bound keeps the initialization stable. A rank-$r$ gate adds $O(r\,(d_h+d))$ parameters per sine layer; for our width-256 networks the total overhead is $1.4\%$ (rank 1), $2.3\%$ (rank 2), $4.1\%$ (rank 4), and $7.7\%$ (rank 8) over the base network. FiRe wraps the nonlinearity, so it composes with any periodic activation, converting SIREN and FINER to FiRe-SIREN and FiRe-FINER, respectively.

\paragraph{Initialization.}
We use the \emph{sine-zeros} initialization, i.e., $\mathbf{U}_\uparrow=0$ and $\mathbf{b}_\uparrow=1$, so at initialization $\mathbf{s}_i = 2\sigma(1)\approx1.46$ for every neuron. The gate is functionally constant and FiRe is, at init, a plain periodic network whose effective frequency is scaled by $1.46$.

\subsection{The \textbf{\emph{wscale}} and Parameter Count Control}
\label{sec:control}
A constant gate is observationally a frequency change, so a raw comparison of FiRe to a baseline conflates the gate's \emph{parameterization} with a \emph{frequency} increase. We remove the frequency channel with the \emph{wscale} control, wherein, every sine-layer weight is scaled by $1/1.46$ at init, so the realized per-neuron frequency $\omega_0\,\mathbf{s}_i\,\lVert W_i\rVert = \omega_0\,\lVert W_i^{\text{base}}\rVert$ equals a plain $\omega_0\!=\!30$ baseline \emph{in every layer} (verified
numerically at init). The $1.46\times$ boost is fully cancelled. Throughout, the comparison baseline is the \textbf{parameter-matched baseline}, i.e., plain SIREN/FINER are \emph{widened} so its parameter count is at least that of FiRe at each rank ($r1\!\to\!w258,\ r4\!\to\!w262$). Every reported $\Delta$ is therefore measured at \emph{matched parameter count and matched realized frequency}. It isolates the gate's optimization-geometry effect.

\subsection{Neural Tangent Kernel Decomposition}
\label{sec:ntk}
We analyze the mechanism through the empirical neural tangent kernel $K(x,x')=\nabla_\theta f(x)^{\!\top}\nabla_\theta f(x') =\sum_{p}\partial_{\theta_p}f(x)\,\partial_{\theta_p}f(x')$, dividing the parameters into the carrier (signal) path $\theta_c=\{\mathbf{W}^{(l)},\mathbf{b}^{(l)}\}$ and the gate path $\theta_g=\{\mathbf{U}_\downarrow^{(l)},\mathbf{U}_\uparrow^{(l)},\mathbf{b}_\uparrow^{(l)}\}$. Because the sum runs over individual parameters and the two sets are disjoint, the kernel is \emph{exactly additive with no cross terms}:
\begin{equation}
\begin{aligned}
K 
&= K_{\text{c}} + K_{\text{g}} \\
&= \!\!\sum_{p\in\theta_c}\!\partial_p f(x)\,\partial_p f(x')
  + \!\!\sum_{p\in\theta_g}\!\partial_p f(x)\,\partial_p f(x')
\end{aligned}
\label{eq:decomp}
\end{equation}
On a finite input set the Gram matrices $K,K_c,K_g=JJ^{\!\top},
J_cJ_c^{\!\top},J_gJ_g^{\!\top}$ satisfy $K=K_c+K_g$ with $K_c,K_g\succeq0$ (Appendix ~\ref{app:ntk}).

\paragraph{Explicit blocks.}
Writing $\mathbf{z}_i=(\mathbf{W}\mathbf{x}+\mathbf{b})_i$ for the pre-activation and $\mathbf{\phi}_i=\cos(\omega_0 \mathbf{s}_i \mathbf{z}_i)$,
a single FiRe unit has carrier and gate gradients
\begin{align}
\frac{\partial \mathbf{h}_i}{\partial \mathbf{W}_{ij}} &= \phi_i\,\omega_0\,\mathbf{s}_i\,\mathbf{x}_j, \label{eq:gradW}\\[2pt]
\frac{\partial \mathbf{h}_i}{\partial (\mathbf{U}_\uparrow)_{ik}} &=
\phi_i\,\omega_0\,\mathbf{z}_i\,\bigl(2\,\sigma'(\cdot)\bigr)\,(\mathbf{U}_\downarrow \mathbf{x})_k . \label{eq:gradU}
\end{align}
The above equations leads to two consequences:

\emph{(i) The gate block is nonzero even though the gate is inert at init.} At initialization $\mathbf{U}_\uparrow=0$, so $\mathbf{s}_i$ is the constant at $1.46$; yet from \eqref{eq:gradU} the \emph{gradient} is $\partial \mathbf{s}_i/\partial(\mathbf{U}_\uparrow)_{ik}=2\sigma'(\mathbf{b}_\uparrow)\,(\mathbf{U}_\downarrow \mathbf{x})_k\neq0$. The gate therefore contributes tangent directions without changing the forward map. By \eqref{eq:gradU} these directions are modulated by the pre-activation $\mathbf{z}_i$ itself (not by the coordinate $\mathbf{x}$ as in \eqref{eq:gradW}), so they carry signal-correlated structure of a different functional form than the carrier directions.

\emph{(ii) Adding $K_{\text{g}}$ can only raise eigenvalues.} Since $K=K_c+K_g\succeq K_c$ in the Loewner order, Weyl's inequality gives $\lambda_i(K)\geq\lambda_i(K_c)$ for all $i$, and $\operatorname{tr}K=\operatorname{tr}K_c+\operatorname{tr}K_g>\operatorname{tr}K_c$. At the \emph{wscale} init the forward map equals the parameter-matched baseline, so $K_c$ is that baseline's kernel up to the gradient rescaling induced by moving the $1.46$ factor from $\mathbf{W}$ into the activation argument (a conditioning effect on \emph{existing} directions; cf.\ the $\omega_0 \mathbf{s}_i$ factor in \eqref{eq:gradW}), while $K_g$ supplies \emph{new} directions (Appendix ~\ref{app:ntk}).

\paragraph{From effective rank to convergence rate.} In the lazy/NTK regime, linearized training makes the residual decay along each kernel eigendirection at a rate set by its eigenvalue: the squared error contributed by eigenvector $\mathbf{v}_i$ decays as $e^{-2\lambda_i t}$, so the training loss obeys $\mathcal{L}(t)\approx\sum_i \pi_i\,e^{-2\lambda_i t}$ for residual projections $\pi_i$. A kernel with more sizable eigenvalues, higher trace and higher effective rank, reduces more error directions quickly, accelerating \emph{early} optimization. This is the bridge between the init-time enlargement we measure (\S\ref{sec:results-ntk}) and the faster convergence we observe (\S\ref{sec:long}); it is an early-time statement, and as training leaves the lazy regime the prediction weakens, consistent with the late-training crossover in Fig.~\ref{fig:long}. The link to implicit acceleration of overparameterization are given in Appendix~\ref{app:ntk}.

\paragraph{What this does and does not establish.} Above sections establish that the gate cannot lower NTK eigenvalues and strictly raises the trace. They do \emph{not} by themselves guarantee a higher \emph{effective rank} (entropy of the normalized spectrum), which would fail if the added mass aligned with the carrier's leading eigenvectors. Empirically it does not: by \eqref{eq:gradU} the gate directions read the input through a separate path and land largely off the carrier's principal axes, raising the effective rank by $23$--$44\%$ (\S\ref{sec:results-ntk}, Table~\ref{tab:ntk}). We therefore treat the eigenvalue/trace statement as theory and the effective-rank increase as the empirically supported mechanism. We emphasize that the mechanism is tangent-space \emph{enlargement}, not condition-number reduction. The acceleration of \S\ref{sec:ntk} follows from raising eigenvalues (trace and effective rank), not from lowering $\lambda_{\max}/\lambda_{\min}$. Though, we observed improvement in condition number as well at rank 4 (Table~\ref{tab:ntk}). The added block $K_g$ contributes off-axis mass that need not touch the spectral endpoints. Effective rank, which integrates over the whole spectrum rather than its two extremes, is therefore the stable correlate of the gain.

\section{Results and Discussions}
We evaluate FiRe on 2D image representation using 16 images from DIV2K~\cite{agustsson2017ntire} at resolutions $256^2$, $512^2$, and $1024^2$, under full-batch MSE training with Adam ($\mathrm{lr}=10^{-4}$, 2000 epochs unless stated otherwise), and we replicate key findings on a second dataset (Kodak -24 images; \S\ref{sec:kodak}). All comparisons are conducted between FiRe (\emph{wscale}) and a parameter-matched plain baseline with $\omega_0=30$ (\S\ref{sec:control}). Paired $\Delta$PSNR (FiRe $-$ baseline; Peak Signal-to-Noise Ratio) is computed for each image and evaluated using the Wilcoxon signed-rank test ($^{*}p<0.05$, $^{**}p<0.01$, $^{***}p<0.001$). For multi-seed experiments, we use $5$ seeds $\times$ $16$ images $=80$ paired fits. Unless noted, PSNR is the best value reached during training (a running maximum); the convergence study (\S\ref{sec:long}) instead reports the instantaneous value at each checkpoint. All baselines are reproduced from their official implementations under identical training settings and matched parameter counts, with hyperparameters following the original papers whenever possible. Additional implementation details are provided in Appendix~\ref{app:impl}.

\subsection{Seed-Robustness of the Parameter-Matched Gain}
\label{sec:robust}
At $256^2$ the gate yields $+0.5$ to $+0.7$\,dB over the parameter-matched baseline, significant across ranks for both activations (Table~\ref{tab:rank}). Because the comparison is paired, what matters is the paired standard error: over $n{=}80$ fits the $95\%$ confidence intervals are narrow and exclude zero for every rank except FINER\,r8 (the one n.s.\ cell). The effect is thus small in magnitude but well separated from noise, with $p$ between $10^{-6}$ and $10^{-11}$.

\begin{table}[t]
\centering\small
\caption{Seed-robustness. Paired $\Delta$PSNR (dB), FiRe $-$ parameter-matched
baseline, $256^2$, $2000$ epochs, $5$ seeds $\times\,16$ imgs ($n{=}80$). Superscripts indicate significance under the \emph{Wilcoxon signed-rank test} ($^{*}p<0.05$, $^{**}p<0.01$, $^{***}p<0.001$; n.s. = not significant).}
\label{tab:rank}
\begin{tabular}{c r r r r}
\toprule
 & rank 1 & rank 2 & rank 4 & rank 8 \\
\midrule
FiRe-SIREN & $+0.64^{***}$ & $+0.70^{***}$ & $+0.72^{***}$ & $+0.49^{***}$ \\
FiRe-FINER & $+0.68^{***}$ & $+0.48^{***}$ & $+0.50^{***}$ & $+0.11^{\text{n.s.}}$ \\
\bottomrule
\end{tabular}
\end{table}

We observed two structural patterns. First, the gain is roughly flat across ranks 1--4 and only drops at rank 8: as the gate widens, the parameter-matched baseline is widened to keep pace, and at rank 8 the baseline is wide enough to absorb the gate entirely (FINER\,r8 is the sole non-significant cell). The effect therefore lives in the low-rank regime where the gate adds a small, coherent preconditioning mode rather than competing capacity. Second, the gain is comparable across configurations despite their different baselines. FINER already applies an input-dependent $(|z|{+}1)$ scaling, yet the gate still helps it by a similar margin, indicating that the preconditioning effect is not specific to the plain sinusoid.

\subsection{Dependence on Resolution}
\label{sec:resolution}
\begin{figure}[t]
\centering
\includegraphics[width=0.8\textwidth]{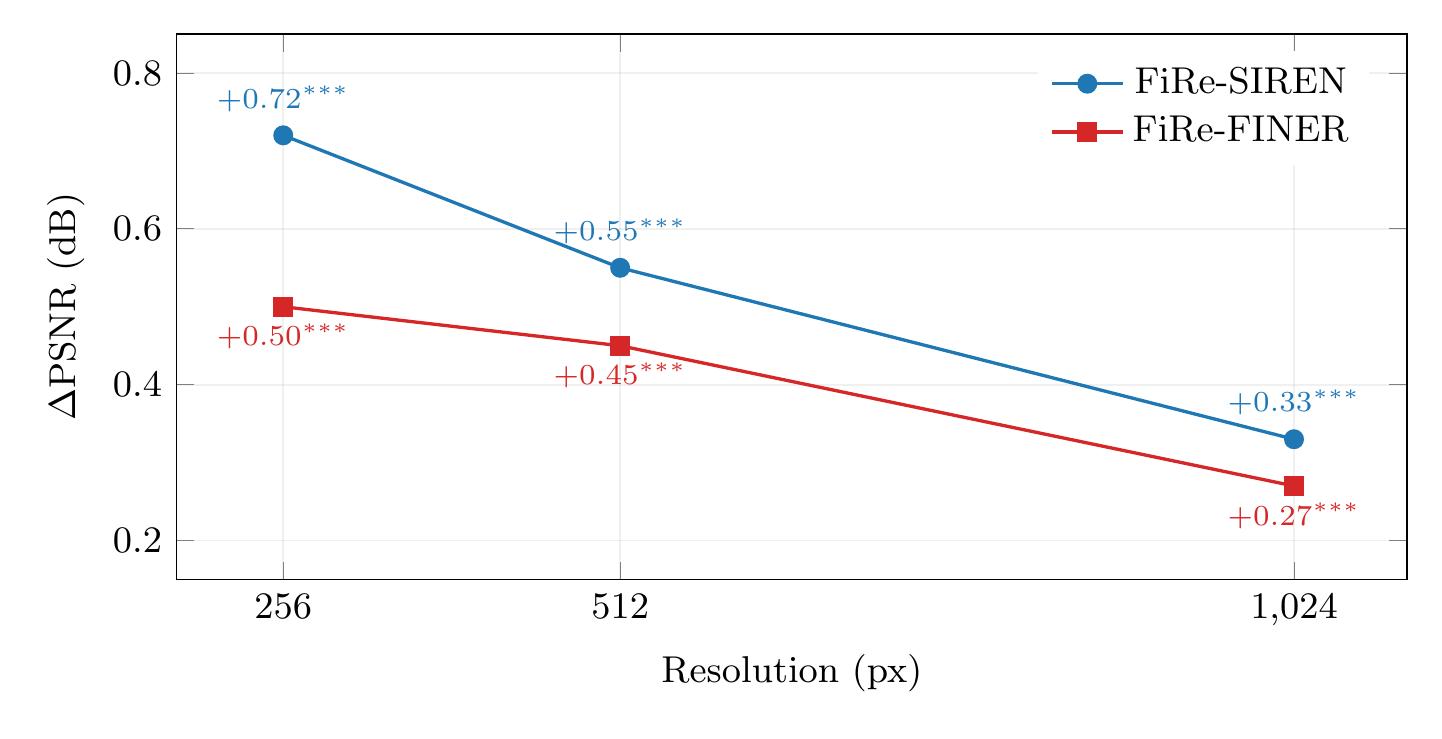}
\caption{The parameter-matched gain vs. resolution (rank 4, $2000$ ep, 5 seeds, DIV2K--16 images). The gain shrinks with resolution, consistent with an optimization-limited effect that attenuates as the fit becomes capacity-limited.}
\label{fig:res}
\end{figure}

At fixed rank~4 the gain declines monotonically with resolution but stays positive and significant everywhere (Fig.~\ref{fig:res}). The decline is substantial, roughly halving from $256^2$ to $1024^2$ ($+0.72\!\to\!+0.33$ for SIREN, $+0.50\!\to\!+0.27$ for FINER) and is consistent across both activations. Since $\Delta$PSNR is a log-ratio of mean-squared errors, a
constant multiplicative convergence advantage would appear as a roughly constant
$\Delta$ across resolution. The observed shrink instead indicates that the
advantage itself attenuates as the problem becomes harder relative to a fixed
budget. The next subsection shows this is the optimization-versus-budget axis seen
from a single training length: higher resolution sits further from convergence at
$2000$ ep, so the gate's head start is a smaller fraction of the remaining climb.

\subsection{Convergence Dynamics Across Training Budgets}
\label{sec:long}
\begin{figure}[t]
\centering
\includegraphics[width=0.8\textwidth]{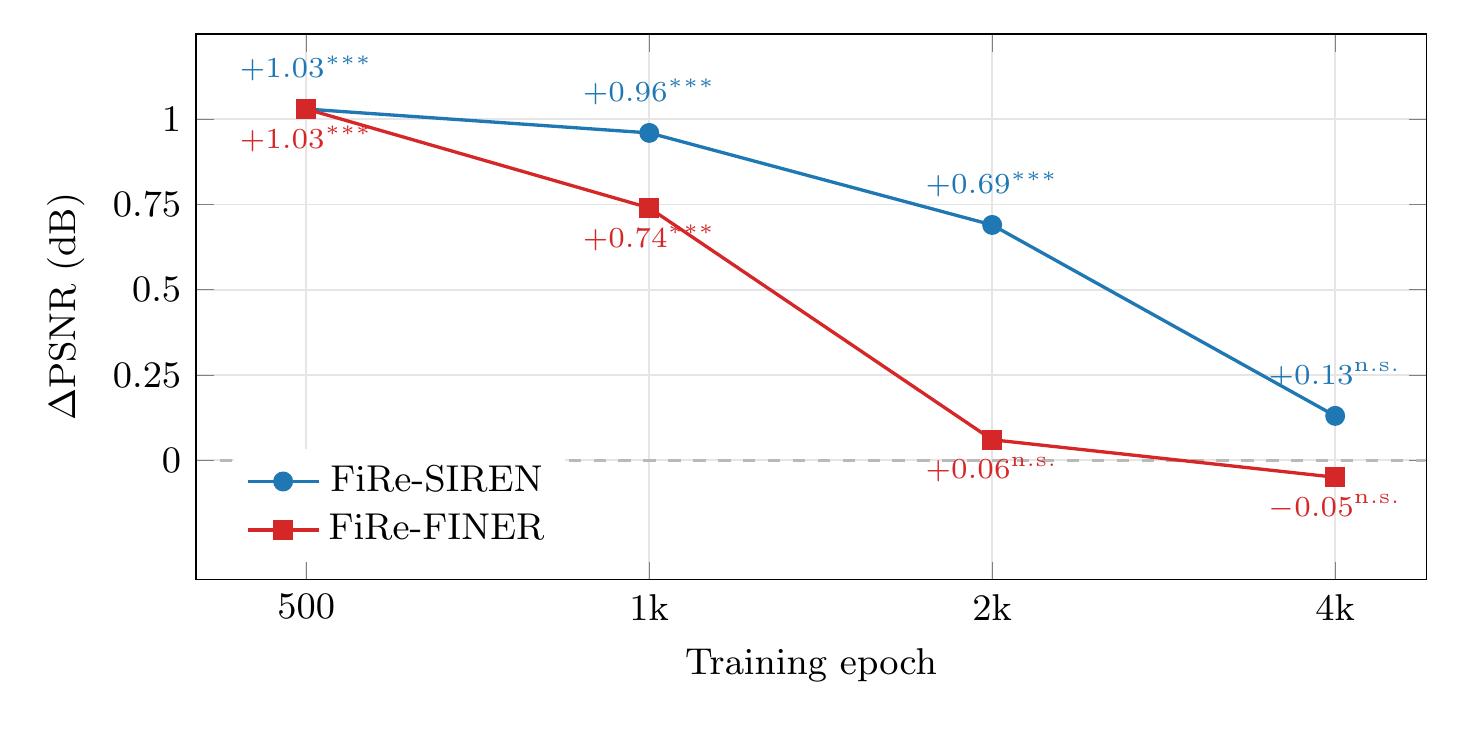}
\caption{Paired $\Delta$PSNR vs. epoch at $256^2$ resolution, rank 4, 5 seeds for DIV2K--16 images. FiRe leads strongly at the short budgets standard in INR fitting and the gap closes toward zero as training proceeds. The advantage is therefore one of preconditioning; a capacity gain would persist.}
\label{fig:long}
\end{figure}

We train FiRe vs.\ the parameter-matched baseline at $256^2$, rank~4, and track the paired gap over training (Fig~\ref{fig:long}; $5$ seeds). The advantage is largest early ($\sim\!+1$\,dB at $500$ ep, $p\!\approx\!10^{-13}$) and decays as training proceeds. The gate reaches a given accuracy in fewer iterations. The two models cross zero at different times. FiRe-FINER's paired gain stays significant until $\sim\!1500$ epochs and SIREN's until $\sim\!2700$ epochs. The parameter-matched baseline overtakes FiRe (at $8000$ ep, $-1.24$ for SIREN and $-0.84$ for FINER), confirming that FiRe holds no asymptotic advantage. The same reversal reproduces on Kodak ($-0.36$ SIREN, $-0.26$ FINER at $8000$ ep, \S\ref{sec:kodak}), so it is not a DIV2K artifact. This is consistent with the unchanged function class (\S\ref{sec:ntk}) confirming that the benefit is an early-optimization effect, exactly as the additive NTK picture predicts, i.e., the gate reshapes the tangent space at initialization, governing early-time dynamics. The prediction is expected to lapse once training leaves the lazy regime, precisely
where the crossover occurs.

\paragraph{Best-PSNR vs. instantaneous PSNR.} The Table~\ref{tab:rank} and Fig.~\ref{fig:res} report different statistics from runs that coincide up to $2000$ ep. Table~\ref{tab:rank} reports the \emph{best PSNR reached during training}, a running maximum over $[0,2000]$, the standard INR metric, whereas the convergence study (\S\ref{sec:long}) reports the \emph{instantaneous} PSNR at each checkpoint. These
diverge once a run passes its peak and begins to overfit. For FiRe-FINER (r4, $256^2$) the best-PSNR advantage is $+0.50^{***}$, but the instantaneous gap at $2000$ ep is only $+0.06$ (n.s.). The best-PSNR gap thus records the early advantage; the instantaneous gap records what remains after FiRe's peak has passed. The discrepancy is a corollary of the acceleration claim.

\paragraph{Resolution and budget are the same axis.} At $1024^2$, even $2\times$ the budget ($4000$ ep) still shows a persisting gap (FiRe-SIREN $+0.18$, FiRe-FINER $+0.28$, vs.\ $+0.22/+0.20$ at $2000$ ep). High resolution is simply further from convergence, so the curves have not yet met. This is why the gain shrinks with resolution at a fixed budget (\S\ref{sec:resolution}). The resolution and training length are two views of the same optimization-versus-budget axis.

\subsection{Qualitative Reconstruction and Derivative Fidelity}
\label{sec:qual}
\begin{figure*}[t]
\centering
\includegraphics[width=\textwidth]{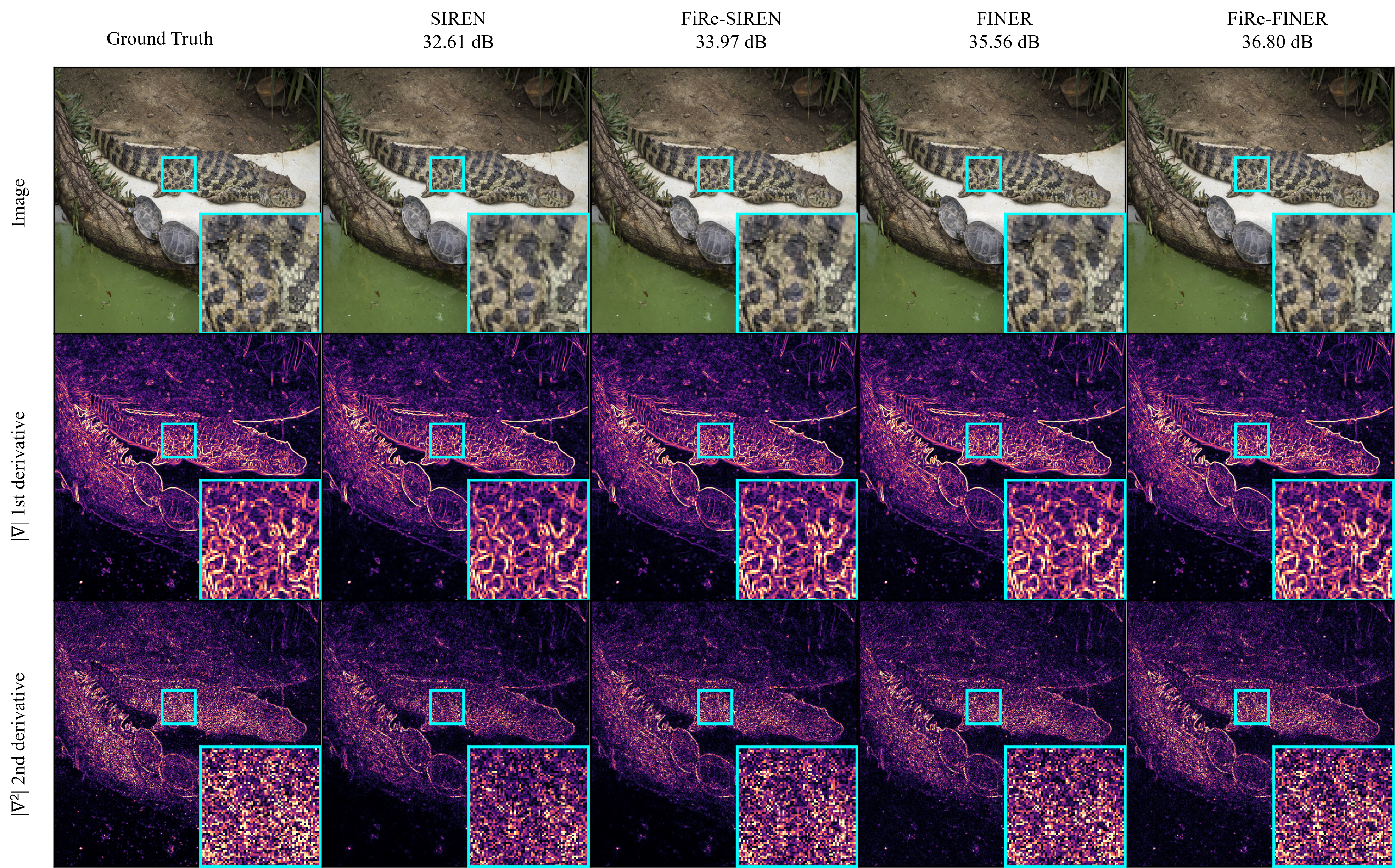}
\caption{Reconstruction and derivative comparison on a DIV2K image. Rows: Image, First derivative, Second derivative; Columns: Ground Truth, SIREN, FiRe-SIREN, FINER, FiRe-FINER. Insets zoom a high-frequency region; differences are clearest in the second derivative. PSNR is per-image.}
\label{fig:qual1}
\end{figure*}

Fig.~\ref{fig:qual1} compare reconstructions and their first
and second spatial derivatives at the standard budget on a DIV2K dataset image. The
derivative panels are the informative view. Because periodic INRs are prized for representing a signal and its derivatives, errors that are subtle in the image become visible in first ($|\nabla|$) and second derivatives ($|\nabla^2|$). Consistent with the spectral analysis (\S\ref{sec:results-ntk}), the difference between FiRe and the baseline is most apparent in the second derivative. In the insets, FiRe recovers finer high-frequency derivative structure, edges and texture, closer to ground truth for both the functions. The figure is illustrative of \emph{where} the gate helps.

\begin{table}[t]
\centering
\caption{NTK at initialization (matched realized frequency). Relative changes are w.r.t. the parameter-matched baseline (eff.\ rank: $\uparrow$ better). For condition number ($\downarrow$ better), we report the \emph{conditioning improvement} multiplier $\kappa_{\mathrm{base}}/\kappa_{\mathrm{FiRe}}$ ($>1$ better).}
\label{tab:ntk}
\begin{tabular}{lccccc}
\toprule
 &  & \multicolumn{2}{c}{eff.\ rank ($\uparrow$)} & \multicolumn{2}{c}{cond. ($\kappa$) ($\downarrow$)} \\
\cmidrule(lr){3-4}\cmidrule(lr){5-6}
backbone & rank & FiRe & param-match & FiRe & param-match \\
\midrule
SIREN & 1 & $138$\,\textcolor{green!50!black}{($\uparrow$23\%)} & $112$ & $2.3\mathrm{e}9$\,\textcolor{red!70!black}{($\times 0.16$)} & $3.7\mathrm{e}8$ \\
FINER & 1 & $423$\,\textcolor{green!50!black}{($\uparrow$30\%)} & $326$ & $1.6\mathrm{e}6$\,\textcolor{green!50!black}{($\times 1.06$)} & $1.7\mathrm{e}6$ \\
\midrule
SIREN & 4 & $141$\,\textcolor{green!50!black}{($\uparrow$44\%)} & $\phantom{0}98$ & $5.8\mathrm{e}9$\,\textcolor{green!50!black}{($\times 22.4$)} & $1.3\mathrm{e}11$ \\
FINER & 4 & $417$\,\textcolor{green!50!black}{($\uparrow$26\%)} & $330$ & $2.5\mathrm{e}6$\,\textcolor{green!50!black}{($\times 2.36$)} & $5.9\mathrm{e}6$ \\
\bottomrule
\end{tabular}
\end{table}

\subsection{Neural Tangent Kernel at Initialization}
\label{sec:results-ntk}
\begin{figure}[t]
\centering
\includegraphics[width=0.85\textwidth]{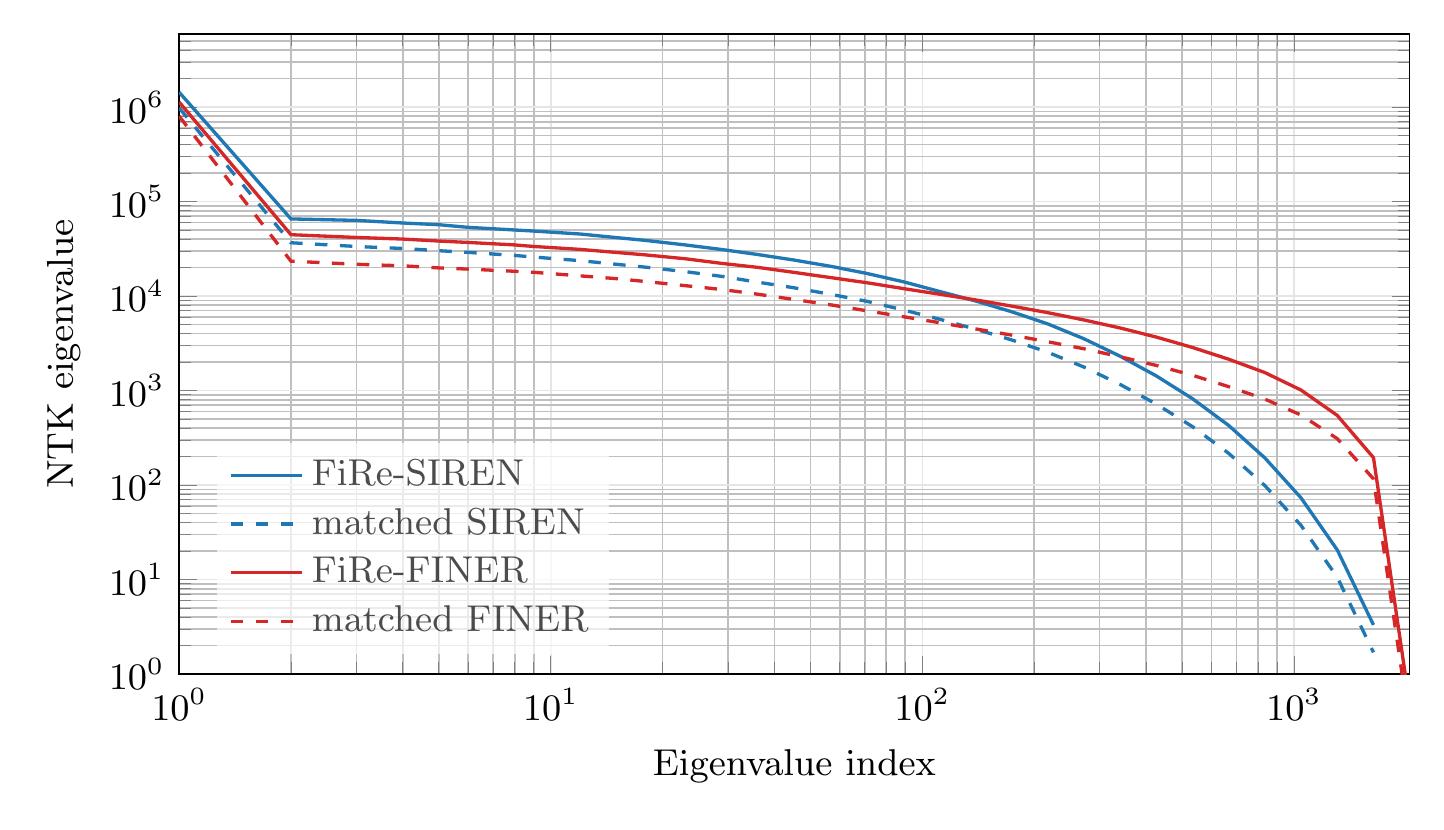}
\caption{NTK eigenvalue decay at initialization (rank 4, $\omega_0{=}30$, $N{=}2048$, $512$px grid, 5-seed mean). At matched effective frequency the FiRe (solid) decays more slowly than the parameter-matched baseline (dashed) across the spectrum for both functions, This is the visual form of the $+23$--$44\%$ effective-rank increase and the mechanistic correlate of the faster convergence.}
\label{fig:eigdecay}
\end{figure}

At matched realized frequency ($\omega_0\!=\!30$, $N{=}2048$ samples, effective rank $=\exp(\text{entropy of eigenvalues})$), the gate enlarges the tangent space in every configuration and effective rank rises by $23$--$44\%$ (Table~\ref{tab:ntk}). The full eigenspectrum decays more slowly for FiRe than for the parameter-matched baseline across the whole index range (Fig.~\ref{fig:eigdecay}), with no eigenvalue dropping below the baseline. This is the spectral statement of Proposition~\ref{prop:weyl} (Appendix~\ref{app:ntk}). 

Two observations further support this. The absolute effective rank differs sharply by function. FINER's baseline kernel already has high effective rank ($\sim\!330$ vs.\ SIREN's $\sim\!100$), reflecting that its $(|z|{+}1)$ scaling itself enlarges the tangent space. However, FiRe still adds a comparable \emph{relative} increase on top, so the enlargement is not redundant with FINER's mechanism. Additionaly, condition number improves markedly for rank 4 (Table~\ref{tab:ntk} but worsens for SIREN\, rank 1). We therefore report effective rank, which moves consistently with the convergence-acceleration gain across every cell, as the robust mechanistic signal.

\begin{figure}[t]
\centering
\includegraphics[width=\textwidth]{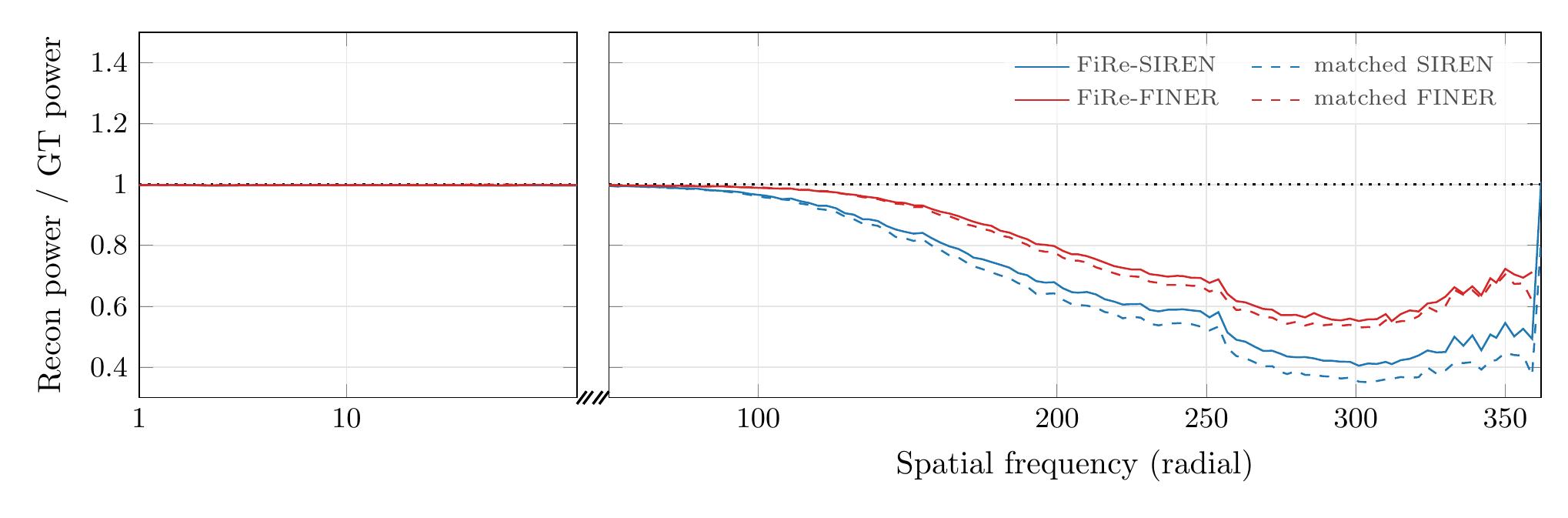}
\caption{High-frequency fidelity: ratio of reconstruction to GT radial power ($512$px, mean over $16$ images $\times\,5$ seeds; ratio${\to}1$ matches GT). The x-axis is broken: a compressed low-frequency stub (ratio $\approx1$) then the high-frequency band $50$--$362$ shown wide. Solid${=}$FiRe, dashed${=}$parameter-matched plain; FiRe retains more high-frequency power than the matched baseline, both families.}
\label{fig:specratio}
\end{figure}

The effective-rank and convergence results predict that the gate's advantage is concentrated in the high-frequency, optimization-limited band. We test this directly by the radially-binned ratio of reconstructed to ground-truth spectral power (Fig.~\ref{fig:specratio}). A ratio of $1$ means the reconstruction matches the target at that spatial frequency. At low and mid frequencies all configurations sit at $1$, i.e., every model fits the easy band. Whereas, in the high-frequency tail FiRe tracks the target more closer to $1$ than the parameter-matched baseline for both functions, recovering high-frequency content the matched baseline is still missing at this budget. Here the diagnostic is proximity to 1, not absolute power.

The same picture accounts for three earlier observations. The second-derivative panels of \S\ref{sec:qual} (which up-weight exactly this band), the resolution shrink (\S\ref{sec:resolution}), and the early-budget crossover (\S\ref{sec:long}) are all the same effect: a preconditioning advantage localized to the frequencies that are slowest to optimize.

\subsection{Realized-Frequency Stability During Training}
\label{sec:drift}
\begin{figure}[t]
\centering
\includegraphics[width=\textwidth]{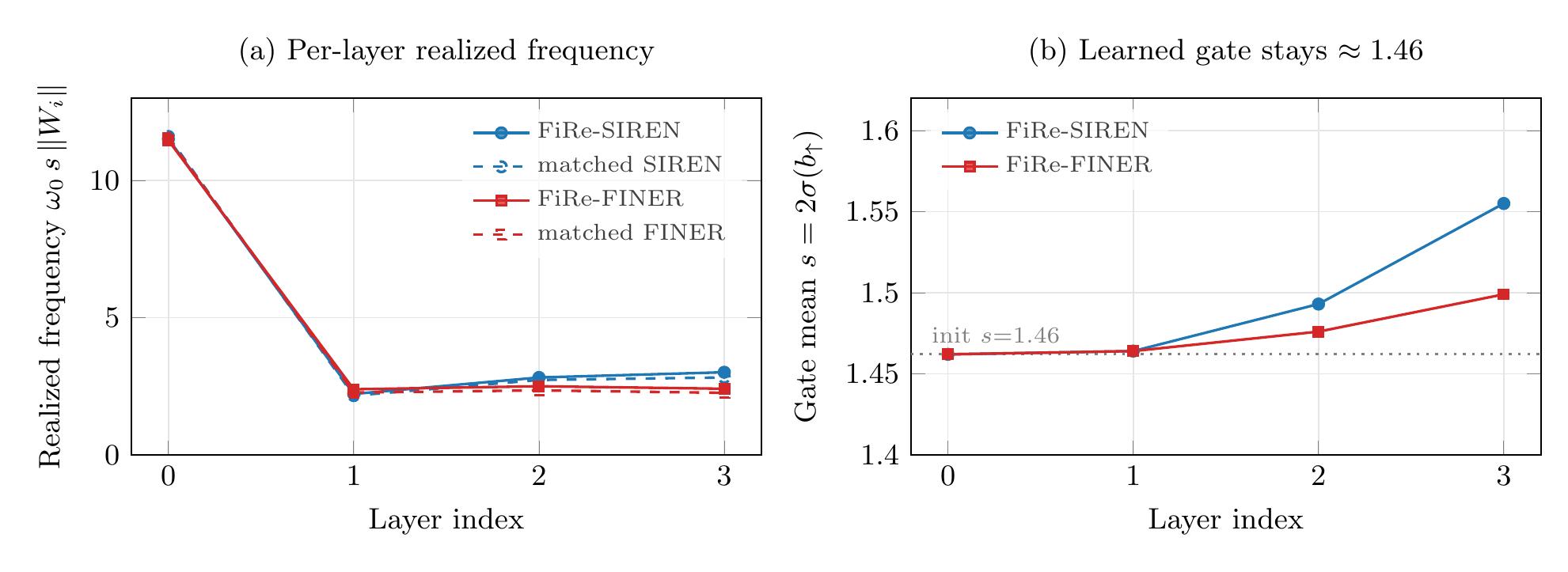}
\caption{\textbf{FiRe operates at matched frequency.} Per-layer statistics at the end of training, mean over $16$ images $\times\,5$ seeds (DIV2K, $512$px, rank~4). \textbf{(a)} The realized per-neuron frequency $\omega_0\,\mathbf{s}\,\lVert \mathbf{W}_i\rVert$ of FiRe (solid) and baseline (dashed) layer by layer. \textbf{(b)} The learned gate mean $\mathbf{s}=2\sigma(\mathbf{s}_{\uparrow})$.}
\label{fig:freq-gate}
\end{figure}

The preconditioning account requires that FiRe and the parameter-matched baseline keep the \emph{same} realized frequency throughout, otherwise the gain could be a hidden frequency change. We log per-layer realized frequency $\omega_0\,\mathbf{s}\,\lVert \mathbf{W}_i\rVert$ and the gate mean $\mathbf{s}$ over training (Fig.~\ref{fig:freq-gate}). FiRe's per-layer realized frequency tracks the parameter-matched baseline and does not separate, and $\mathbf{s}_i$ stays $\approx\!1.46$ (drifting at most to $\approx\!1.55$ in deep layers). The gain at matched frequency thus reflects reparameterization rather than a hidden frequency change. Layer-wise drift over training are additionally shown in Fig.~\ref{fig:freq-gate-dist} and Fig.~\ref{fig:freq-dist-iter-siren} (Appendix~\ref{app:extra}).

\subsection{Replication on a Second Dataset (Kodak)}
\label{sec:kodak}
The 2D findings replicate on Kodak (24 images, rank 4, 5 seeds;
Table~\ref{tab:kodak}). At the standard budget the gate gives $+1.10$\,dB
(FiRe-SIREN) and $+0.99$\,dB (FiRe-FINER) at $256^2$, shrinking to $+0.79$ and $+0.66$ at
$512^2$, with $p$ down to $10^{-20}$ over $120$ paired fits. On an independent
dataset with larger $n$, it follows the same sign, rank-4 magnitude band, and resolution shrink seen on DIV2K dataset. Trained to $8000$ ep the parameter-matched baseline
overtakes ($-0.36$, $-0.26$; both significant), so the convergence-acceleration
reading and the absence of an asymptotic advantage are dataset-independent. The
initialization NTK is computed on the coordinate grid alone and is therefore
identical to DIV2K's (Table~\ref{tab:ntk}); Kodak adds an independent replication
of the \emph{behavioural} findings rather than a second NTK measurement. Per-layer
realized frequency again tracks the parameter-matched baseline throughout, with
slightly larger late-training drift than on DIV2K. The qualitative comparison, effective rank improvement and spectral power ratio on Kodak dataset is shown in Table~\ref{tab:kodakomega}, Fig.~\ref{fig:qual2} and Fig.~\ref{fig:spectral-ratio-kodak-broken} (Appendix~\ref{app:extra})

\begin{table}[t]
\centering\small
\caption{Paired $\Delta$PSNR (dB) showing FiRe $-$ parameter-matched replication on Kodak dataset (24 images, rank 4, 2000 eps, $5$ seeds, at two resolutions). Superscripts indicate significance under the \emph{Wilcoxon signed-rank test} ($^{***}p<0.001$)}
\label{tab:kodak}
\begin{tabular}{l c c c}
\toprule
 & $256^2$ & $512^2$ \\
\midrule
FiRe-SIREN & $+1.10^{***}$ & $+0.79^{***}$ \\
FiRe-FINER & $+0.99^{***}$ & $+0.66^{***}$ \\
\bottomrule
\end{tabular}
\end{table}

\section{Limitations}
\label{sec:limitations}
FiRe accelerates convergence at a fixed training budget, it does not change the converged solution. The gain is largest early in training and at the short budgets typical of INR fitting. Because the gate leaves the function class unchanged (\S\ref{sec:ntk}) it gives no advantage at full convergence, where the parameter-matched baseline matches or exceeds FiRe (\S\ref{sec:long}). The benefit appears when optimization is the bottleneck (low resolution, rank up to 4) and disappears when capacity binds (FINER rank 8). The NTK account is an initialization-time argument, the eigenvalue and trace statements are exact, but the effective-rank increase and its link to convergence rate rest on empirical measurement and the lazy-regime approximation. Finally, our experiments cover 2D image fitting only.

\section{Conclusion}
\label{sec:conclusion}
We introduced FiRe, a per-neuron frequency gate for periodic INRs, and isolated its mechanism by holding realized frequency and parameter count fixed. Under that control the gate's separate parameter path adds a positive-semidefinite block to the neural tangent kernel, raising its effective rank at initialization and speeding up early optimization without changing the function class. The benefit is a convergence acceleration which is spectrally localized to the high-frequency band, largest at the short budgets typical of INR fitting, and gone at full convergence.

The result also carries a methodological warning. A constant-gate reparameterization looks like a frequency change, and a wider network looks like added capacity, so any apparent gain should be checked against frequency- and parameter-matched controls before being credited to a new mechanism. In practice FiRe helps most where INRs are trained to a fixed, short budget rather than to convergence. Open questions include how the trade-off behaves in deployment, whether the realized-frequency control extends to SDF and NeRF, and whether the preconditioning view can be used to design the gate rather than only to explain it.


{\small
\bibliographystyle{ieee_fullname}
\bibliography{references}
}

\appendix
\onecolumn
\section{Extended Neural Tangent Kernel Analysis}
\label{app:ntk}
 
\subsection{Setup and notation}
For a network output $f_\theta(x)\in\mathbb{R}$ and parameters $\theta$, the
empirical NTK on inputs $X=\{x_1,\dots,x_N\}$ is the Gram matrix $K=JJ^{\!\top}$,
$K_{mn}=\langle\nabla_\theta f(x_m),\nabla_\theta f(x_n)\rangle$, where $J$ is the
$N\times|\theta|$ Jacobian. FiRe partitions $\theta=\theta_c\cup\theta_g$ into
carrier and gate parameters, giving $J=[J_c\,|\,J_g]$ and, since the blocks share
no columns, $K=J_cJ_c^{\!\top}+J_gJ_g^{\!\top}=K_c+K_g$ with $K_c,K_g\succeq0$.
 
\subsection{Why the gate enlarges the kernel}
From \eqref{eq:gradW}--\eqref{eq:gradU}, the carrier columns for unit $i$ are
proportional to $\phi_i\,\mathbf{x}$ and the gate columns to $\phi_i\,\mathbf{z}_i\,(\mathbf{U}_\downarrow
\mathbf{x})$, with $\phi_i=\cos(\omega_0 \mathbf{s}_i \mathbf{z}_i)$. The two families differ in their
coordinate dependence: carrier columns are linear in the input $\mathbf{x}$, gate columns
are linear in the \emph{pre-activation} $\mathbf{z}_i=\mathbf{W}_i \mathbf{x}$ passed through the low-rank
map $\mathbf{U}_\downarrow$. Generic position of these two function classes means $K_g$
places mass in directions only partially aligned with the top eigenvectors of
$K_c$; entropy of the normalized spectrum (effective rank) then increases. This is
the empirically observed $+23$--$44\%$ (Table~\ref{tab:ntk}, Fig.~\ref{fig:eigdecay}).
We state the bound that \emph{is} guaranteed:
\begin{proposition}[Monotone spectrum]
\label{prop:weyl}
For any FiRe layer, $K\succeq K_c$, hence $\lambda_i(K)\ge\lambda_i(K_c)$ for all
$i$ and $\operatorname{tr}K=\operatorname{tr}K_c+\operatorname{tr}K_g$ with
$\operatorname{tr}K_g>0$ whenever $\mathbf{U}_\downarrow \mathbf{x}\neq0$ for some training input.
\end{proposition}
\noindent Proposition~\ref{prop:weyl} follows from $K_g\succeq0$ and Weyl's
inequality. It guarantees no eigenvalue decreases and the trace strictly
increases; it does \emph{not} guarantee higher effective rank, which is the
empirical content of \S\ref{sec:results-ntk}.
 
\subsection{From kernel mass to early-training speed}
Under gradient flow in the lazy regime, $f$ evolves approximately linearly and the
residual $r(t)=f_t(X)-y$ obeys $\dot r=-K\,r$, so in the eigenbasis $\{(\lambda_i,
v_i)\}$ of $K$,
\[
\mathcal{L}(t)=\tfrac12\|r(t)\|^2=\tfrac12\sum_i \langle r(0),v_i\rangle^2\,
e^{-2\lambda_i t}.
\]
Larger eigenvalues drain their error directions faster; raising the trace and
spreading mass over more sizable eigenvalues (Proposition~\ref{prop:weyl} plus the
empirical effective-rank increase) lowers $\mathcal{L}(t)$ faster at small $t$.
This predicts an \emph{early}-training advantage that need not persist: the lazy
approximation degrades as $\theta$ moves, and once it does the kernel argument no
longer governs the dynamics. The late-training crossover in Fig.~\ref{fig:long} is
consistent with this boundary of validity, and with the function class being
unchanged so that both parameterizations share the same minima.
 
\subsection{Relation to implicit acceleration of overparameterization}
The multiplicative form $\omega_0\,\mathbf{s}\,\mathbf{W}$ is a depth-two factorization of the
realized frequency channel. Such factorizations are known to precondition gradient
descent without changing the represented function. We give the minimal scalar
instance. Let a single realized frequency $\Omega$ be fit under a loss
$\mathcal{L}(\Omega)$, comparing a direct parameterization $\Omega=c$ to a
factored one $\Omega=a\,b$ (the gate $a$ times the weight $b$). Under gradient
flow,
\[
\text{direct:}\ \dot\Omega=-\mathcal{L}'(\Omega);\qquad
\text{factored:}\ \dot a=-\mathcal{L}'(\Omega)\,b,\ \ \dot b=-\mathcal{L}'(\Omega)\,a,
\]
so that $\dot\Omega=\dot a\,b+a\,\dot b=-(a^2+b^2)\,\mathcal{L}'(\Omega)$.
\begin{remark}
The factorization multiplies the effective learning rate on $\Omega$ by
$(a^2+b^2)$, a parameter-dependent preconditioner $>1$ for the FiRe init
($a=s\approx1.46$, $\mathbf{b}=\lVert \mathbf{W}\rVert$). It accelerates the approach to the same
stationary point. It does not move the stationary point, mirroring the empirical
finding that FiRe speeds early convergence without changing the converged
solution. The full gate is low-rank and coordinate-dependent, so the
network-level effect is the anisotropic, off-axis enlargement of $K_g$ above
rather than a single scalar factor; this scalar case is the mechanism in
miniature.
\end{remark}
 
\section{Implementation and Reproducibility}
\label{app:impl}
\noindent \textbf{Architecture:} Width-256 MLPs, 3 hidden sine layers, $\omega_0=30$, output linear. Baselines use the standard SIREN initialization; FINER adds the $(|\mathbf{z}|{+}1)$ activation. FiRe adds the rank-$r$ gate of Eq.~\eqref{eq:fire} with \emph{sine-zeros} init ($\mathbf{U}_\uparrow{=}0$, $\mathbf{b}_\uparrow{=}1$) and the \emph{wscale} weight rescaling (\S\ref{sec:control}).

\noindent \textbf{Parameter matching:} Parameter-matched baselines are widened to $\ge$ FiRe's count at each rank ($r1{\to}w258$, $r2{\to}w259$, $r4{\to}w262$, $r8{\to}w266$); FiRe param counts (2D) are $201.7$k\,/\,$203.5$k\,/\,$207.1$k\,/\,$214.3$k for $r1$--$r8$.

\noindent \textbf{Training:} Full-batch MSE, Adam, lr $10^{-4}$; $2000$ epochs (main), extended to $8000$ for the convergence-dynamics study. DIV2K-16 at $256^2/512^2/1024^2$. Kodak-24 at $256^2/512^2$.

\noindent \textbf{NTK computation:} Empirical NTK on $N{=}2048$ sampled coordinates at initialization, $\omega_0{=}30$; effective rank $=\exp$ of the eigenvalue entropy.
 
\section{Additional Empirical Results}
\label{app:extra}
 
\noindent \textbf{Per-seed convergence breakdown}\\
Table~\ref{tab:perseed} reports the paired $\Delta$PSNR for each individual seed ($n{=}16$ images each) at $500$ and $2000$ ep, for the convergence-dynamics study of \S\ref{sec:long} ($256^2$, rank 4). The early-budget advantage is positive for
every seed and both backbones at $500$ ep. By $2000$ ep the per-seed values are more dispersed. FINER in particular shows seed-dependent sign (e.g.\ $-1.33$ for seed~0, $+1.31$ for seed~3) which is the instantaneous-metric behaviour discussed in \S\ref{sec:long}: as the per-image PSNR peaks earlier under FiRe and then plateaus, the instantaneous gap at a fixed late epoch becomes noisy across seeds while the best-PSNR-so-far (Table~\ref{tab:rank} and Fig.~\ref{fig:long}) remains consistently positive.
 
\begin{table}[h]
\centering\small
\caption{Per-seed paired $\Delta$PSNR (dB), FiRe $-$ parameter-matched, $256^2$,
rank 4 ($n{=}16$ per cell). Seeds 2 and 3 are the additions to the original
three-seed run.}
\label{tab:perseed}
\begin{tabular}{l c r r r r r}
\toprule
model & ep & seed-1 & seed-2 & seed-3 & seed-4 & seed-5 \\
\midrule
FiRe-SIREN & 500  & $+0.78$ & $+1.04$ & $+1.02$ & $+0.83$ & $+1.48$ \\
FiRe-FINER & 500  & $+0.71$ & $+1.03$ & $+1.49$ & $+0.82$ & $+1.08$ \\
\midrule
FiRe-SIREN & 2000 & $+1.19$ & $+0.70$ & $+0.86$ & $+0.26$ & $+0.43$ \\
FiRe-FINER & 2000 & $+0.74$ & $-1.33$ & $-0.13$ & $-0.31$ & $+1.31$ \\
\bottomrule
\end{tabular}
\end{table}
 
\noindent \textbf{Per-layer realized frequency (Kodak)}\\
Table~\ref{tab:kodakomega} gives the per-layer realized frequency
$\omega_0\,\mathbf{s}\,\lVert \mathbf{W}_i\rVert$ at initialization and at convergence, for FiRe and
the parameter-matched baseline on Kodak ($512^2$, rank 4). FiRe tracks the
baseline at every layer; the gate mean $\mathbf{s}$ stays near its initial $1.46$, drifting
to at most $1.54$ in the deepest layer. The late-training drift is slightly larger
than on DIV2K but does not separate the two curves, supporting the
matched-frequency (reparameterization) reading of \S\ref{sec:drift}.
 
\begin{table}[h]
\centering\small
\caption{Kodak per-layer realized frequency $\omega_0\,\mathbf{s}\,\lVert \mathbf{W}_i\rVert$
(init$\to$final) and gate mean $\mathbf{s}$, rank 4, $512^2$ ($120$ runs each).}
\label{tab:kodakomega}
\begin{tabular}{c l l c}
\toprule
layer & FiRe eff-$\omega$ & param-match eff-$\omega$ & FiRe gate $s$ \\
\midrule
\multicolumn{4}{l}{\emph{SIREN}}\\
0 & $11.40\!\to\!11.45$ & $11.50\!\to\!11.55$ & $1.46\!\to\!1.46$ \\
1 & $1.41\!\to\!2.18$ & $1.41\!\to\!2.12$ & $1.46\!\to\!1.46$ \\
2 & $1.42\!\to\!2.68$ & $1.42\!\to\!2.58$ & $1.46\!\to\!1.49$ \\
3 & $1.42\!\to\!2.83$ & $1.42\!\to\!2.64$ & $1.46\!\to\!1.54$ \\
\midrule
\multicolumn{4}{l}{\emph{FINER}}\\
0 & $11.40\!\to\!11.40$ & $11.50\!\to\!11.50$ & $1.46\!\to\!1.46$ \\
1 & $1.41\!\to\!2.31$ & $1.41\!\to\!2.20$ & $1.46\!\to\!1.46$ \\
2 & $1.42\!\to\!2.40$ & $1.42\!\to\!2.26$ & $1.46\!\to\!1.48$ \\
3 & $1.42\!\to\!2.32$ & $1.42\!\to\!2.17$ & $1.46\!\to\!1.50$ \\
\bottomrule
\end{tabular}
\end{table}
 

\begin{figure}[H]
\centering
\includegraphics[width=\textwidth]{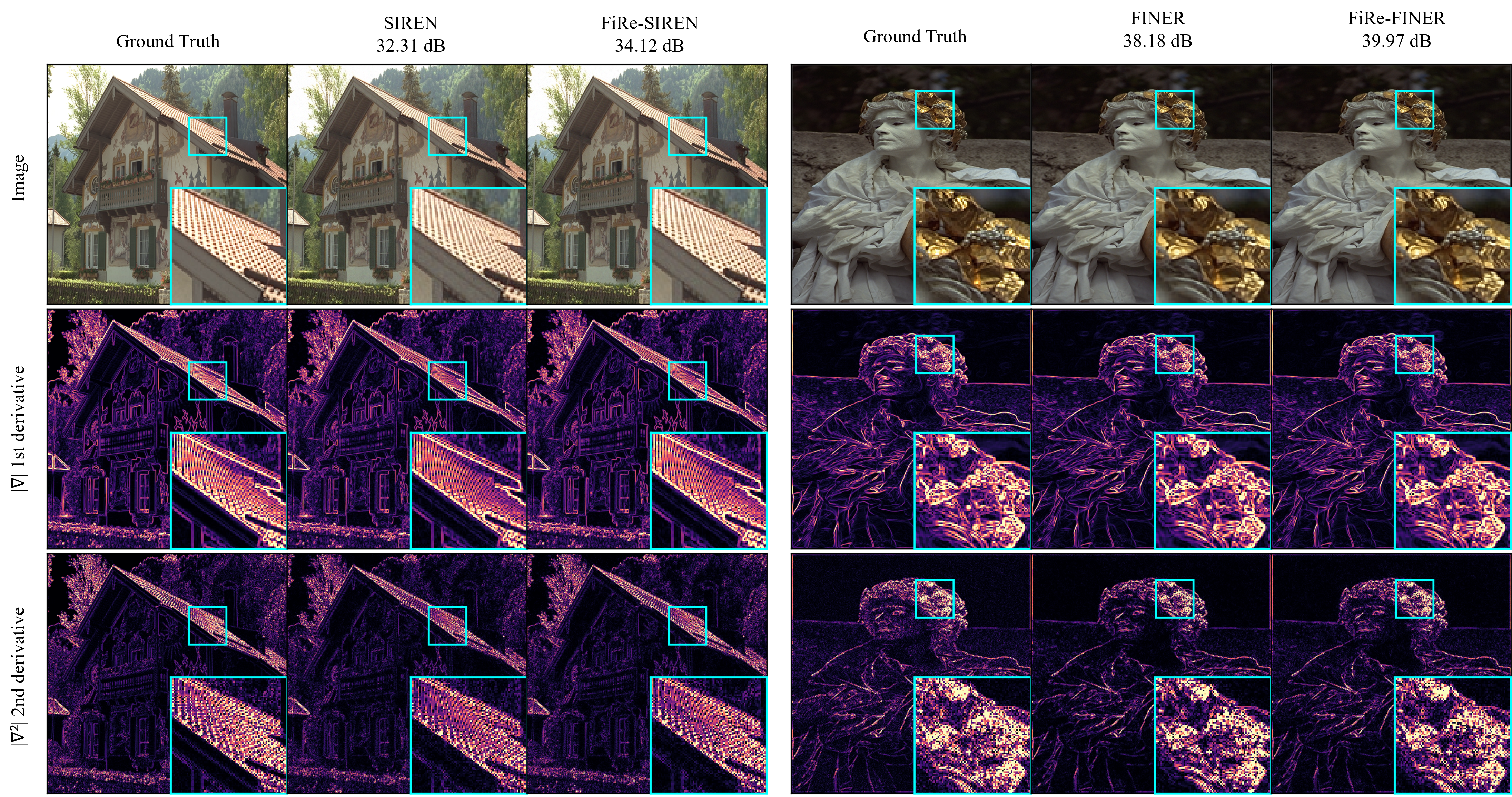}
\caption{Reconstruction and derivative comparison on a Kodak images. Rows: Image, First derivative, Second derivative; Insets zoom a high-frequency region; differences are clearest in the second derivative. PSNR is per-image.}
\label{fig:qual2}
\end{figure}

\begin{figure}[t]
\centering
\includegraphics[width=\textwidth]{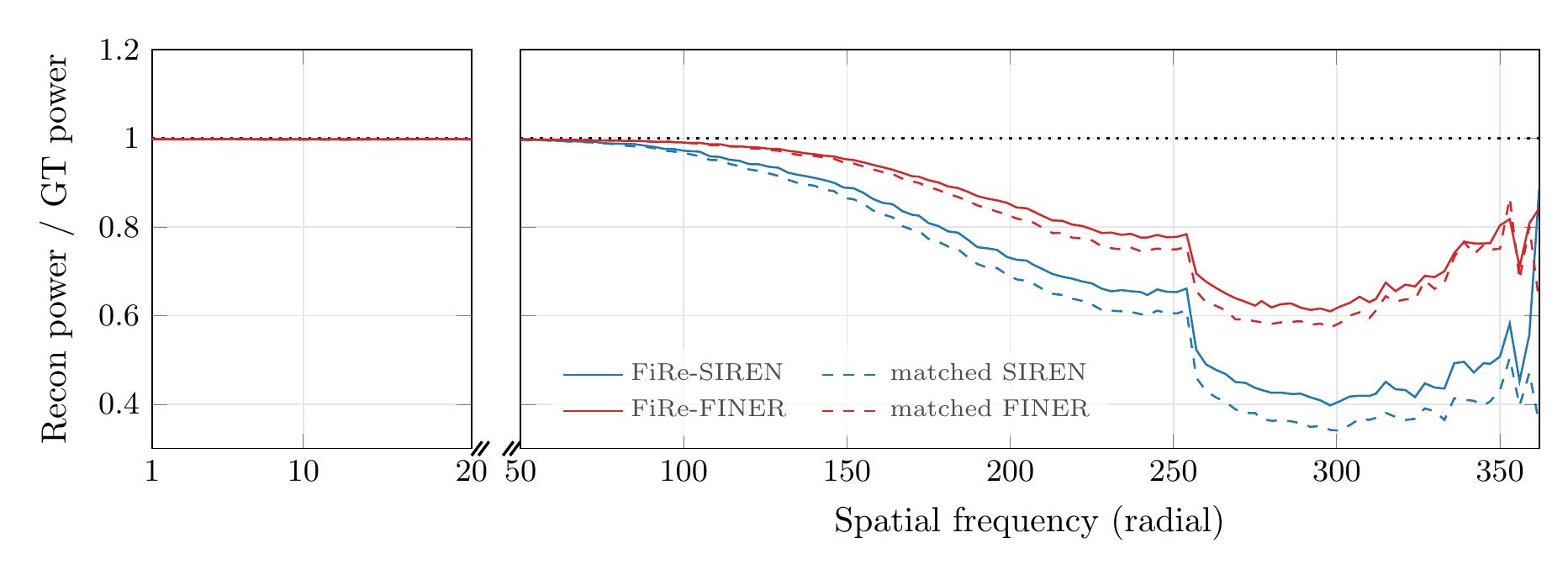}
\caption{High-frequency fidelity on \textbf{Kodak}: ratio of reconstruction to GT radial power
($512$px, mean over $24$ images $\times\,5$ seeds; ratio${\to}1$ matches GT). The x-axis is broken: a
compressed low-frequency stub (ratio $\approx1$) then the high-frequency band $50$--$362$ shown wide.
Solid${=}$FiRe (b1\_wscale), dashed${=}$parameter-matched plain; FiRe retains more high-frequency
power than the matched baseline for both families.}
\label{fig:spectral-ratio-kodak-broken}
\end{figure}

\begin{figure}[t]
\centering
\includegraphics[width=\textwidth]{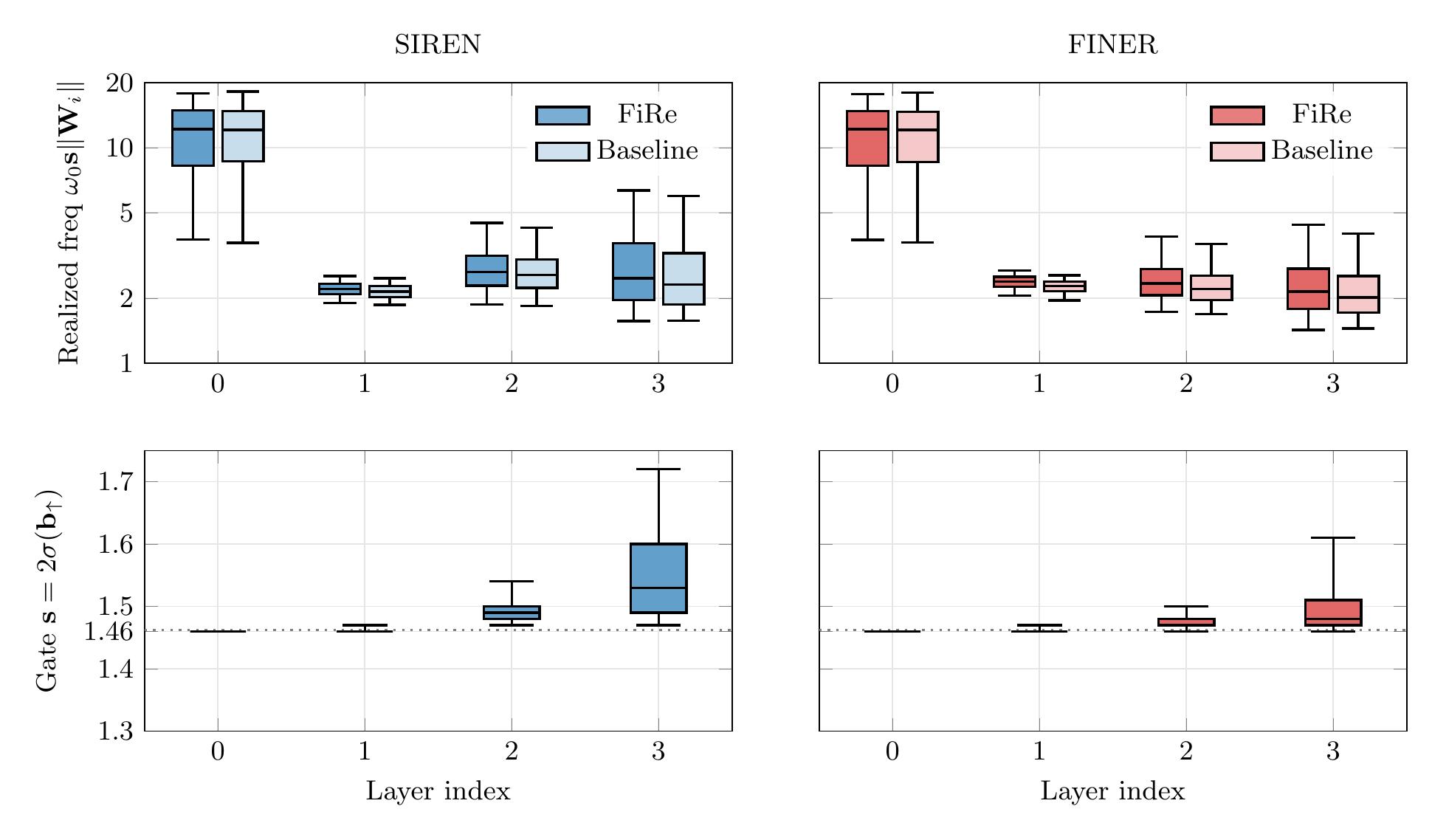}
\caption{\textbf{Distribution of per-neuron realized frequency and learned gate.} Boxplots pooled over all neurons $\times\,16$
images $\times$ all seeds (DIV2K, $512$px, rank~4). \textbf{Top:} realized per-neuron frequency
$\omega_0 \mathbf{s}\lVert \mathbf{W}_i\rVert$ (log scale); FiRe and parameter-matched baseline share the same distribution at every layer, for both families. \textbf{Bottom:} the learned gate
$\mathbf{s}=2\sigma(\mathbf{b}_\uparrow)$ stays tightly concentrated about its initialization $\approx1.46$ (dotted). The gain is therefore a matched-frequency reparameterization effect, not a frequency change.}
\label{fig:freq-gate-dist}
\end{figure}

\begin{figure}[t]
\centering
\includegraphics[width=\textwidth]{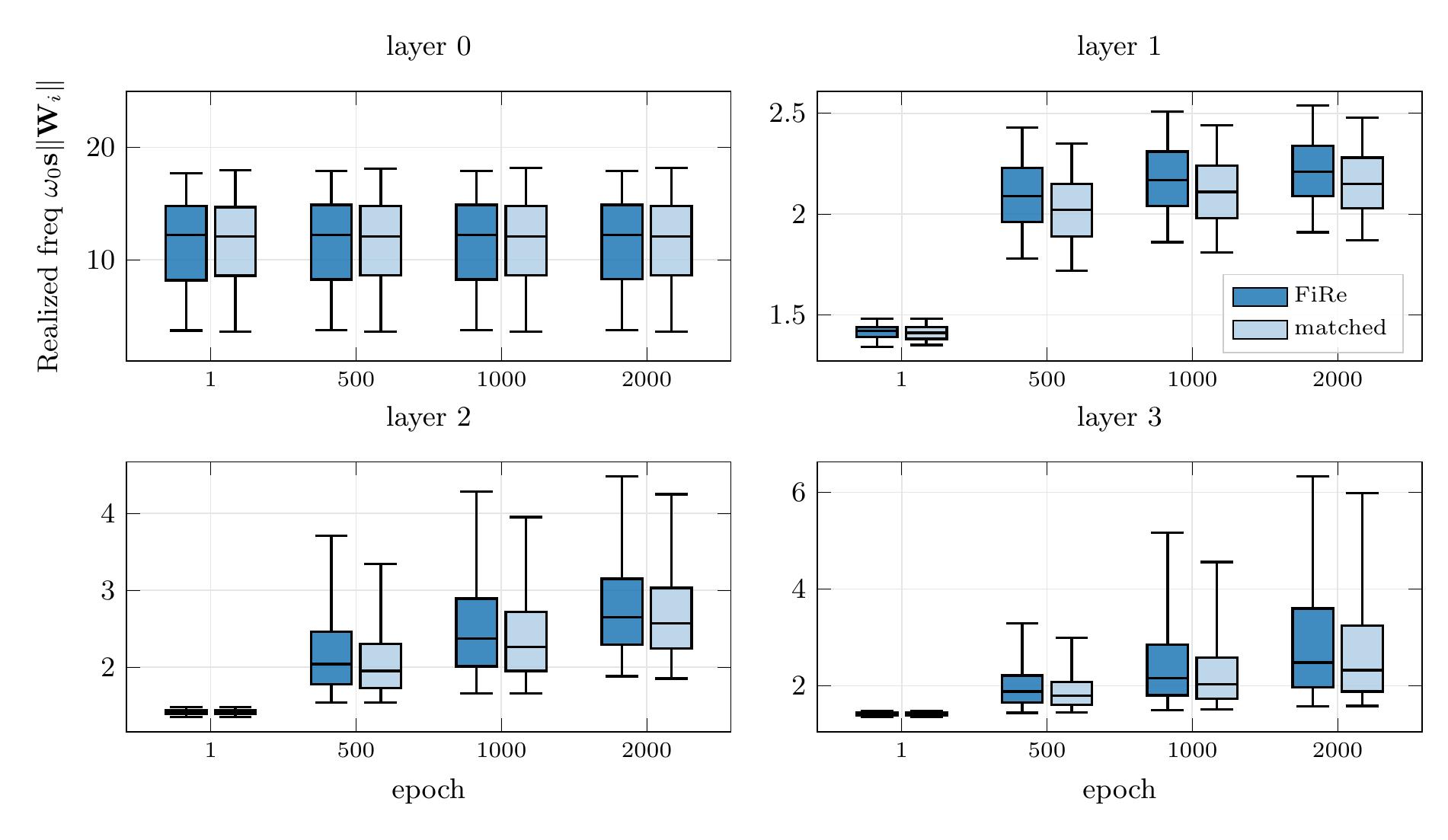}
\caption{\textbf{SIREN (rank~4): per-neuron realized frequency $\omega_0 \mathbf{s}\lVert \mathbf{W}_i\rVert$ distribution over training.} Boxplots pooled over all neurons $\times\,16$ images $\times$ all seeds (DIV2K, $512$px), at epochs $\{1,500,1000,2000\}$. FiRe and parameter-matched baseline evolve through the \emph{same} frequency distribution at every layer and every checkpoint, i.e., the gate induces no frequency drift relative to the matched model.}
\label{fig:freq-dist-iter-siren}
\end{figure}

\end{document}